\pdfoutput=1

\documentclass[11pt]{article}

\usepackage[]{EMNLP2023}

\usepackage{times}
\usepackage{latexsym}

\usepackage[T1]{fontenc}

\usepackage[utf8]{inputenc}

\usepackage{microtype}

\usepackage{inconsolata}

\usepackage{array}
\usepackage{makecell}

\NewDocumentCommand\githubicon{}{\includegraphics[scale=0.025]{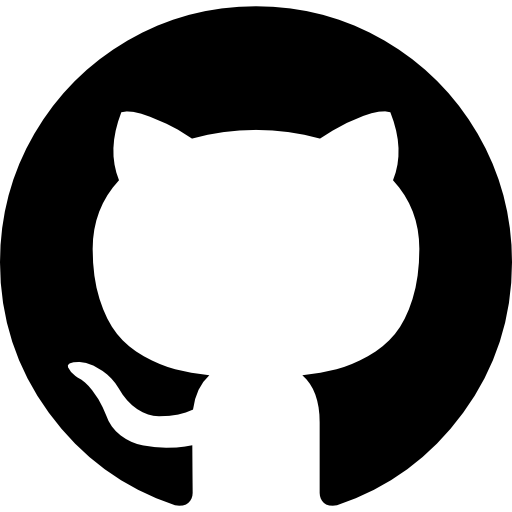}
}

\usepackage{hyperref}
\usepackage{graphicx}
\usepackage{booktabs}
\usepackage{multirow}
\usepackage{enumitem}
\usepackage{amsmath}
\usepackage{todonotes}
\usepackage{graphicx}
\usepackage{caption}
\usepackage{subcaption}
\usepackage{microtype}
\usepackage[nameinlink]{cleveref}
\usepackage{pdfpages}
\usepackage{tabularx}

\everypar{\looseness=-1}

\title{Thorny Roses: Investigating the Dual Use Dilemma\\in Natural Language Processing}

\author{
\parbox{\linewidth}{\centering
Lucie-Aim\'{e}e Kaffee$^{1}$, 
Arnav Arora$^{2}$, 
Zeerak Talat$^{3}$,  
Isabelle Augenstein$^{2}$}\\
\parbox{\linewidth}{\centering
  $^{1}$Hasso Plattner Institute, Germany,
  $^{2}$University of Copenhagen, Denmark \\
  $^{3}$Mohamed Bin Zayed University of Artificial Intelligence, United Arab Emirates\\
  \texttt{lucie-aimee.kaffee@hpi.de,  aar@di.ku.dk, z@zeerak.org,  augenstein@di.ku.dk }
  }
  }

\begin{document}
\maketitle

\begin{abstract}
Dual use, the intentional, harmful reuse of technology and scientific artefacts, is an ill-defined problem within the context of Natural Language Processing (NLP).
As large language models (LLMs) have advanced in their capabilities and become more accessible, the risk of their intentional misuse becomes more prevalent.
To prevent such intentional malicious use, it is necessary for NLP researchers and practitioners to understand and mitigate the risks of their research.
Hence, we present an NLP-specific definition of dual use informed by researchers and practitioners in the field.
Further, we propose a checklist focusing on dual-use in NLP, that can be integrated into existing conference ethics-frameworks. 
The definition and checklist are created based on a survey of NLP researchers and practitioners.\footnote{\githubicon We make the survey and checklist available at: \url{https://github.com/copenlu/dual-use}}
\end{abstract}

\section{Introduction}
As usage of NLP artefacts (e.g., code, data, and models) increases in research and commercial efforts, it becomes more important to examine their rightful use and potential misuse. %
The Association for Computing Machinery %
code of ethics\footnote{\href{https://www.acm.org/code-of-ethics}{Association for Computing Machinery Code of Ethics}}, which has been adopted by ACL\footnote{\href{https://www.aclweb.org/portal/content/acl-code-ethics}{Association for Computational Linguistics Code of Ethics}}, states that \textit{''[c]omputing professionals should consider whether the results of their efforts (...) will be used in socially responsible ways''}.
New regulation in the European Union also requires producers of LLMs to outline foreseeable misuses and mitigation techniques~\cite{EU_proposal_2023}.
However, without guidance on how to assess the impacts on society, researchers and practitioners (henceforth professionals) are ill-equipped to consider social responsibility in deciding on tasks to work on or managing resulting research artefacts.
This is reflected in contemporary ethical review processes which %
emphasise the impacts of research on individual subjects, rather than the wider social impacts of conducted research. 

While very few research projects have malicious motivations, some are reused to harm any but particularly marginalised groups of society.
This presents a crucial gap in the ethics of artificial intelligence and NLP on malicious reuse, or \emph{dual use}, which has been particularly absent in literature.

Concerns of dual use of Artificial Intelligence (AI) have been %
discussed by prior work~\cite[e.g.,][]{DBLP:journals/natmi/ShankarZ22,kania2018technological,DBLP:journals/see/SchmidRR22,DBLP:journals/natmi/UrbinaLIE22, ratner2021sweetie, DBLP:conf/tto/GamageSC21}.
However, NLP technologies are rarely included in such considerations.
As LLMs are being incorporated into a wide range of consumer-facing products, the dual use considerations become increasingly critical for research and deployment.
For instance, 
a online mental health services company has %
used ChatGPT to respond to people seeking advice and aid, raising concerns of informed consent, and impact on vulnerable users~\cite{Biron_Online_2023}. 
To address such misuse of LLMs, it is crucial to support researchers in making decisions to limit the dual use of their work.

We propose a checklist that can be implemented by researchers to guide them in the decision making.
If integrated into a paper, the checklist further serves to elucidate the considerations made and approaches taken to mitigate dual use to other professionals.
We keep the checklist concise, such that it can readily be integrated into existing checklists for ethics used by conferences within NLP.
The checklist is developed based on a mixed-methods survey. %
In the survey of NLP professionals, we find that the majority of our respondents do not spend a significant amount of time considering dual use concerns arising from their work and that institutional support often is inadequate.

In summary, our paper contributes: (a) a definition of dual use for NLP artefacts; (b) a survey of NLP professionals documenting their concerns of misuse of NLP artefacts; and (c) a checklist and outlines of methods for mitigating risks of dual use.
\section{Dual Use in NLP}\label{sec:dual-use-nlp}
NLP artefacts, such as LLMs, can have severe impacts on society, even when they are not developed with malicious intent. %
Addressing the issue of malicious use, the European Union defines dual use as: \textit{``goods, software and technology that can be used for both civilian and military applications''}.\footnote{\url{https://policy.trade.ec.europa.eu/help-exporters-and-importers/exporting-dual-use-items_en}} 
However, similarly to many other definitions of dual use, this definition only transfers to a subset of NLP projects. From the field of neuroscience, ~\citet{mahfoud-etal-2018} suggest that the current military use and export control lens for dual use is not adequate for scientists and policy-makers to anticipate harms and recommend investigating the political, security,
intelligence, and military domains of application. Drawing from dual use research in life sciences,~\citet{koplin_dual-use_2023} highlight the need for considerations in AI in light of NLP models as sophisticated text generators. 

We argue that the field of NLP is at a crossroad -- on one hand, artefacts have become more accessible.
On the other, the increased access to resources limits the control over their use.
While availability and open access can facilitate reproducible science, there is a need to develop practices that prevent the malicious use of artefacts.
Discussing the social impacts of NLP, \citet{DBLP:conf/acl/HovyS16} describe dual use as \textit{``unintended consequences of research''}.
However, this definition leaves malicious reuse out of scope.
\citet{DBLP:journals/see/Resnik09} argue that dual use definitions should neither be too narrow or too broad in scope, as that results in limiting what can be considered or make mitigating risks of dual use unmanageable. %
\citet{forge2010note} further posit that the context of dual use is critical in determining its implications and context-specific threats and risks should be specified.

Therefore, in our work, we present a NLP specific definition of dual use, and an associated checklist that is based on the insights from our survey of the field. Our definition is founded in a classification of harms~\cite{DBLP:journals/corr/abs-2112-04359}.

\subsection{Defining Dual Use in NLP}\label{sec:defining}

We conduct a survey of NLP professionals (see \Cref{sec:survey-dual-use-definition}) and found that participants were interested in a precise definition that communicates clearly to professionals who are unfamiliar with dual use.
We therefore formulate a NLP specific definition of dual use that is informed by our survey results and prior work.
Our definition centres the intent of work being created, and the intent with which it is reused.
While our definition is informed by our insights into the NLP research community, it may be applicable to other fields.
We advise that researchers from other fields carefully evaluate whether their field would require any changes to the definition to capture field-specific challenges.

\paragraph{\textbf{Definition}}
\textit{We understand \textbf{dual use} as the \textbf{malicious reuse} of technical and research artefacts that were developed without harmful intent. 
\textbf{Malicious reuse} signifies %
applications that are used to harm any, and particularly marginalised groups in society, where \textbf{harm} describes the perceived negative impacts or consequences for members by those groups.} %
That is, dual use describes the intentional and malicious re-use of for a harmful (secondary) purpose besides its primary application area.
To define harm in the context of dual use, we differentiate between sanction and violence, where our definition of harm focuses on violence rather than sanction.
Briefly, \textit{sanction} can be described simply as violence that is perceived as justified by entities which hold power (e.g., penalties by a government for breaking regulations on technological use), whereas \textit{violence} can simply be described as harms without leave (e.g., breaking regulation which causes harm) \cite{foucault2012discipline,bentham1996collected}. The primary mode through which these differ is by how power is distributed. %

\paragraph{\textbf{Exclusions}}
We exclude from our definition the \textit{unintended secondary} harms from research artefacts.
For instance, our definition does not address the gendered disparities produced in machine translation, as the reproduction of patriarchal values is a secondary harm that arises from the social context and is reflected as model and dataset biases \cite{vanmassenhove2018getting,hovy-etal-2020-sound}.
Similarly, we exclude from our definition tasks where the primary uses are malicious or harmful, e.g. developing research artefacts to predict recidivism rates, %
as the \textit{primary} use-case is harmful~\citep{DBLP:conf/acl/LeinsLB20}.
\section{Survey of NLP Professionals%
}\label{sec:survey}
In order to develop a checklist, we survey
members of the NLP community on their stances %
towards misuse of research artefacts through an anonymous online form.
\citet{DBLP:journals/corr/abs-2208-12852} conduct a survey of NLP professionals' perspectives on general topics, e.g. potential decline in industry interest and promising research directions, dual use is notably absent from consideration.
Indeed, only one question alludes to the issue:
``\textit{Is it unethical to build easily-misusable systems?}'' to which $59\%$ respond in the affirmative, 
indicating that professionals are responsible for minimising misuse within their practice.
We similarly conduct an anonymous survey of NLP professionals, seeking detailed insights into their views on dual use concerns.
Our survey sought to elicit participants understanding of dual use, their perception of harms arising from it, and how to mitigate such harms.

Our adapted survey questions were shaped by feedback on a preliminary version shared researchers within the same group, at different levels of seniority and discussions with colleagues in academia and industry.
The survey was open to the NLP community, widely, from April until the end of June 2022 and advertised on Twitter, professional mailing lists and relevant Reddit communities.\footnote{Reddit communities: \textit{r/nlproc}, \textit{r/MachineLearning}, \textit{r/LanguageTechnology}}
The survey was conducted on the LimeSurvey\footnote{https://www.limesurvey.org/} platform, and approved by the University of Copenhagen ethics board under number 504-0313/22-5000.
The survey contained a total of 23 questions, which consist of a mixture of multiple-choice questions, free-form text answers, and Likert scales (see~\Cref{app:survey-setup} for the survey setup).
We inductively code all free-form text answers (see \Cref{tab:codebook-hamrs} for the codebook).
For our analysis, we discard all partial responses, resulting in $n=48$ complete responses out of $256$ participants. We discarded partial responses as they were mostly limited to the participant demographics questions (the first part of the survey).

\begin{figure}[!tbp]
    \includegraphics[width=\columnwidth]{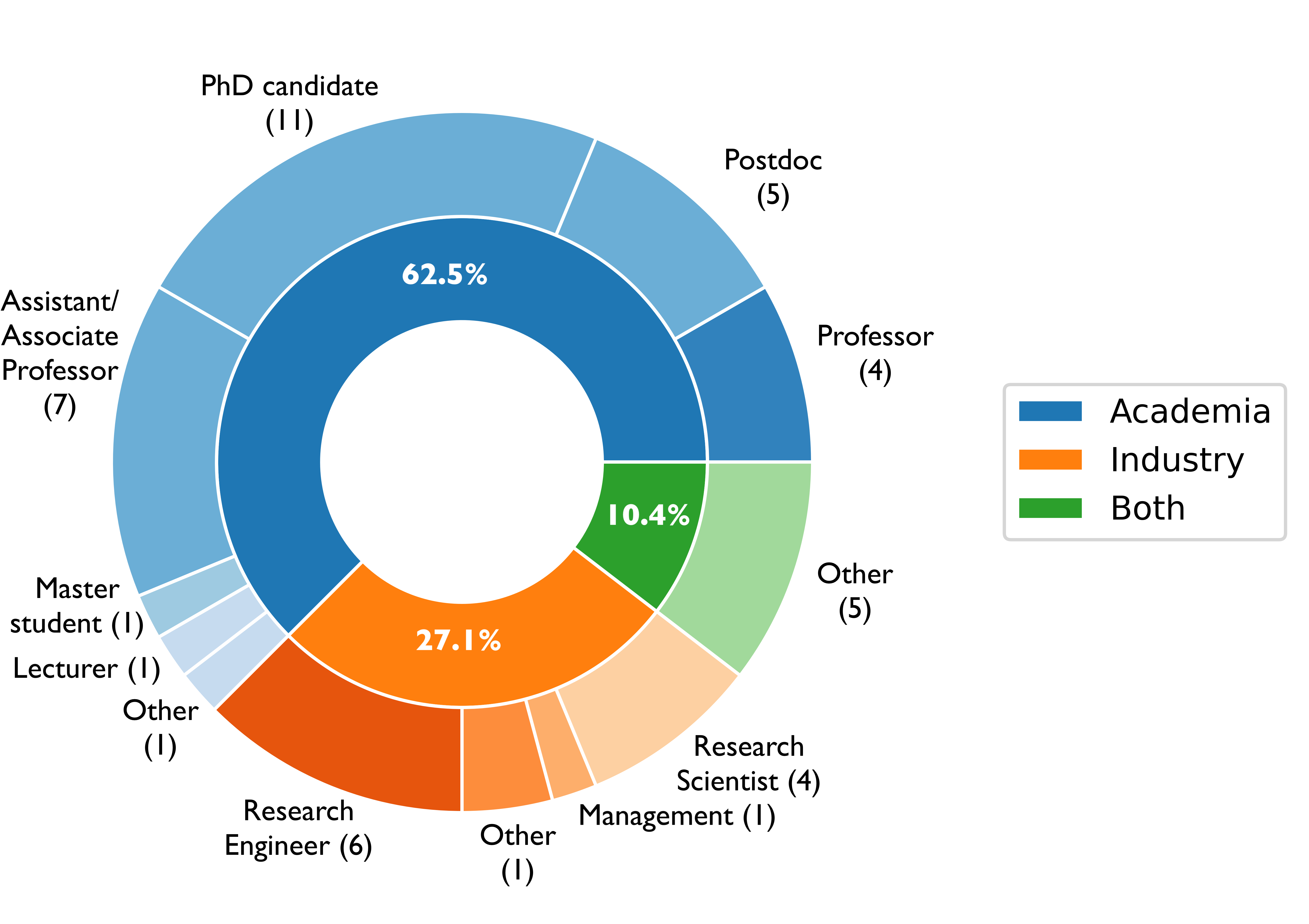} 
  \caption{
  Survey respondents' occupation.
  }\label{fig:occupation}
\end{figure}

\begin{figure}[!t]
    \centering
    \includegraphics[width=\columnwidth]{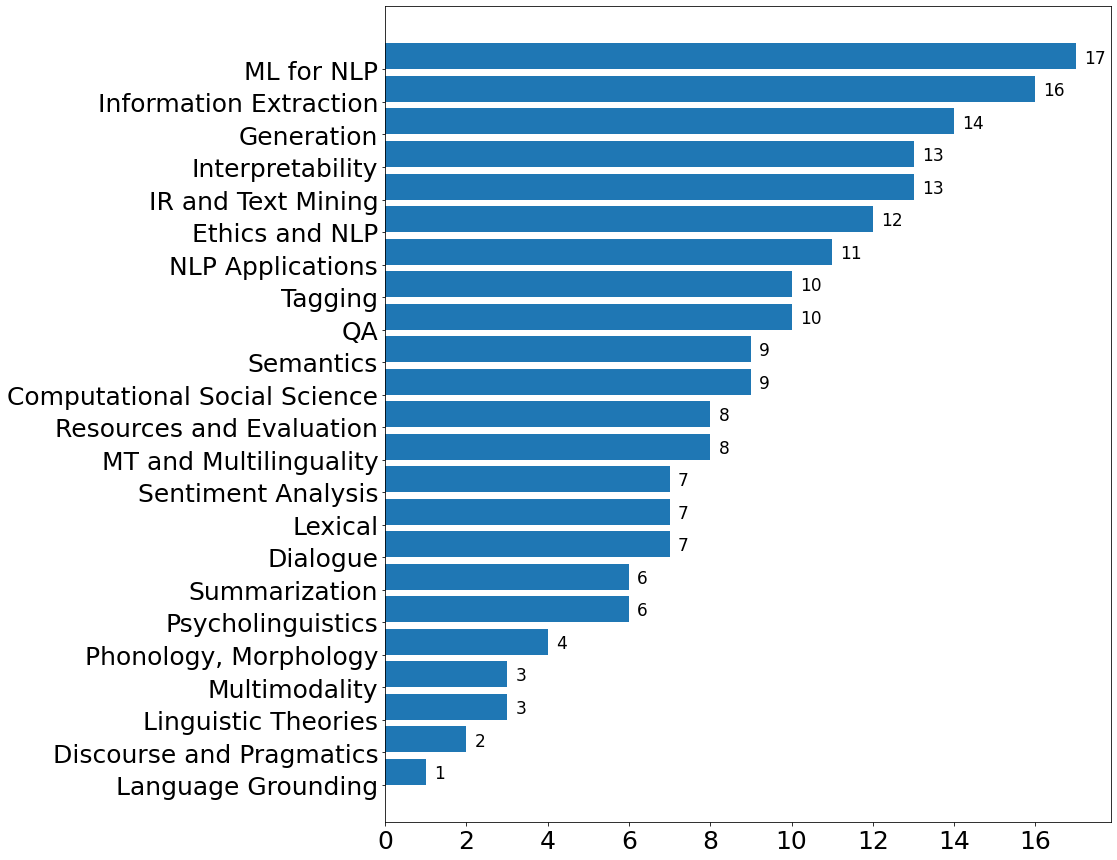}
    \caption{Respondents' area of work.%
    }
    \label{fig:field-bar}
\end{figure}
\subsection{Demographics}\label{sec:survey-demographics}
The majority of participants work in academia (62.5\%) across career stages (see \Cref{fig:occupation}), with PhD students comprising the largest group of participants.
Of participants from industry, most work as Research Scientists.
Participants are primarily based in Europe and North America with only three participants based in Asia, and one in Africa.
This limits the validity of our results to Europe and North America (see \Cref{sec:limitations} for further discussion).

\paragraph{\textbf{Areas of work}}
Participants are asked to select their areas of work and the task they work on, using the ACL 2020 classification of research areas.\footnote{\url{https://acl2020.org/calls/papers/\#submissions}}
Participants can select multiple areas and were asked to add at least one task per area. %
Their responses of work cover a wide range of topics demonstrating a high degree of diversity of areas of work and expertise (see \Cref{fig:field-bar} for the distribution of responses).
This is further reflected by participants working on $180$ unique tasks.
Using participants' input, we identify potential harms for each task in \Cref{sec:survey-perception-harm}.

\subsection{Definition of Dual Use}\label{sec:survey-dual-use-definition}
As developing a NLP specific definition of dual use %
was one of the primary goals of conducting our survey, we presented participants with the following definition of dual use and a free-form text field to respond:
\begin{quote}
    ``Dual use describes any task that can be intentionally used for a harmful (secondary) purpose besides its main application area.''
\end{quote}
While a slight majority of participants (56\%) agree with the definition, %
their responses revealed that a majority found the definition to be too vague.
For instance, participants requested clearer definitions of \textit{``task''} and \textit{``harmful''} such that %
the definition would be clear to people unfamiliar with dual use.
Further, participants emphasised the need to consider the power relations afforded by the definition.
Given the participant responses, we adapt our definition of dual use to the one presented in \Cref{sec:defining}.

\begin{figure*}[t]
    \centering
    \includegraphics[width=\textwidth]{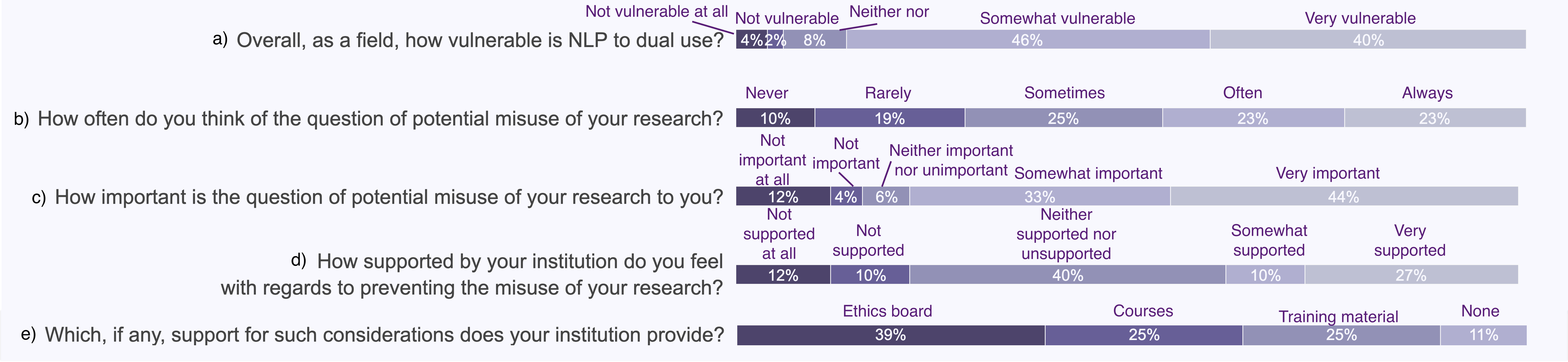}
    \caption{%
    Likert scales of responses 
    on \textit{perceptions of harm} %
    and institutional support for \textit{%
    preventing %
    misuse}. %
    }
    \label{fig:likert-scales}
\end{figure*}
\subsection{Perception of harm}\label{sec:survey-perception-harm}
Given that NLP %
professionals %
shape and implement NLP technology, the survey explores participants' perception of harms in the field of NLP (see \Cref{fig:likert-scales}).
When we asked about the risks of dual use, %
a majority of participants found NLP overall vulnerable, see \Cref{fig:likert-scales}\hyperref[fig:likert-scales]{a}. 
Indeed, 
we find that a majority of participants (77\%) perceive dual use as somewhat or very important to their research, however only a minority (46\%) of participants reported to think about it often or always. %
This demonstrates that while professionals hold %
a general concern, it is not frequently considered in NLP projects.

The survey asks participants how they approach dual use considerations in their research.
The answers cover a broad range of concerns.
For instance, a large share of participants considered how scientific artefacts may be reused:
\textit{[P. 11]: "How could this tool be used for harmful purposes?"}.
Participants also raised questions that indicate impacts on what they work on, e.g., \textit{"Should I build this?"}.
Participants were further concerned about the application area of their work:
\begin{quote}
    [P. 28]: \textit{"How could a corporation or government use this to surveil people?"}
\end{quote}
The wide range of questions and concerns raised by participants provided insight into the nuanced consideration of dual use by NLP professionals and serve as a basis for the checklist (see \Cref{sec:checklist}).
\begin{figure}
    \centering
    \includegraphics[width=\columnwidth]{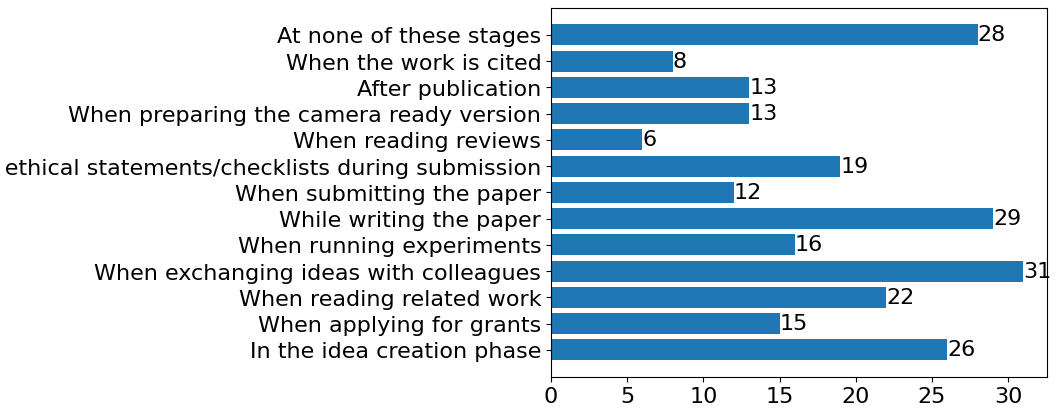}
    \caption{Participant responses to which stage of a project they consider misuse.}
    \label{fig:stages-misuse}
\end{figure}
\paragraph{\textbf{Dual Use in the Research Workflow}}
In efforts to understand the stages in which dual is considered, the participants select stages of their research project in which they consider the risks of misuse (for an overview of responses, see \Cref{fig:stages-misuse}).
Although 58.3\% of respondents selected that they did not find the proposed stages fitting to when they considered misuse, they also selected stages that most closely fit the stage in which they consider misuse.
This indicates that further specificity is required in identifying which stages such considerations occur.
For the respondents who select one or more stages for considering dual use, the discussions predominantly happen in collaborative parts of the research process, i.e. ideation and writing phases.
Less consideration is given after papers have been submitted and in the post publication processes.
While it is important to consider dual use in the early phases, i.e. when selecting tasks to work on, potential harms can often be more clearly identified once scientific artefacts have been produced.
Our checklist (see \Cref{sec:checklist}) therefore seeks to aid professionals in considering dual use in the later stages of research.

\begin{figure}
    \centering\includegraphics[width=0.8\columnwidth]{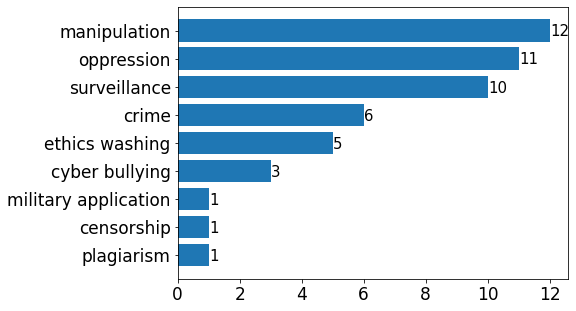}
    \caption{Distribution of codes for harms identified by participants of the tasks in NLP they work on. %
    }
    \label{fig:harms}
\end{figure}

\paragraph{\textbf{Harms by NLP tasks}}\label{sec:task-mapping}
We define the areas of work in the survey using the areas of work described in the ACL 2020 call for papers.\footnote{
\href{https://acl2020.org/calls/papers/\#long-papers}{ACL 2020 Call for Papers.}}
We ask participants to identify the areas and tasks that they work on to obtain expert views on dual use in these fields.
For each task they are asked to provide (1) harmful use cases and (2) how vulnerable they are to misuse.
We then codify the harms described, assigning one code per described harm (see \Cref{fig:harms}).
Intentional manipulation, e.g. dis/misinformation and attempted polarisation of users is frequently listed as a potential harm.
The use language models for such tasks has previously been discussed as \textit{Automated Influence Operations} in detail by~\citet{DBLP:journals/corr/abs-2301-04246}.
Further frequently identified harms include oppressing %
(marginalised) groups of society, e.g. using research artefacts to produce discriminatory outcomes, 
and surveillance by government or corporate entities.
Other concerns centre around reusing NLP technologies for criminal purposes (e.g., fraud), ethics washing, plagiarism, censorship, and military applications.

Participants were also asked to score how vulnerable each task is to misuse on a Likert scale from 1 to 5, where 5 is \textit{very vulnerable} (see \Cref{tab:survey-vulnerability} for aggregate scores and harms associated with each area).
For each area, we find a wide range in harms for associated tasks, thus indicating the importance of considering dual use within specific use-cases.
\begin{table}
    \centering
    \tiny
    \begin{tabular}[\columnwidth]{lcl}
        \toprule
        Area & vulnerability & harms\\
        \midrule
        NLP Applications  & 4.3 & crime, oppression, manipulation\\
        Ethics and NLP & 4.1 & \begin{tabular}[x]{@{}l@{}}ethics washing, surveillance,\\plagiarism, oppression\end{tabular}\\
        Psycholinguistics & 4.0 & cyber bullying, oppression\\
        Generation & 3.8 & \begin{tabular}[x]{@{}l@{}}crime, cyber bullying, manipulation,\\oppression, ethics washing\end{tabular}\\
        Dialogue Systems & 3.8 & surveillance, crime\\
        MT and Multilinguality & 3.3 & surveillance, crime\\
        ML for NLP & 3.2 & \begin{tabular}[x]{@{}l@{}}military application, manipulation,\\oppression\end{tabular}\\
        Resources and Evaluation & 3.1 & \begin{tabular}[x]{@{}l@{}}ethics washing, surveillance,\\manipulation\end{tabular}\\
        Interpretability & 3.0 & ethics washing, manipulation\\
        Information Extraction & 2.8 & surveillance, censorship\\
        IR and Text Mining & 2.7 & surveillance\\
        \bottomrule
    \end{tabular}
    \caption{Average score for vulnerability across ACL areas (with at least three answers) the participants work on and their associated harms. %
    }
    \label{tab:survey-vulnerability}
\end{table}
\subsection{Prevention of Misuse}\label{sec:survey-prevention}
\begin{figure}
    \centering\includegraphics[width=\columnwidth]{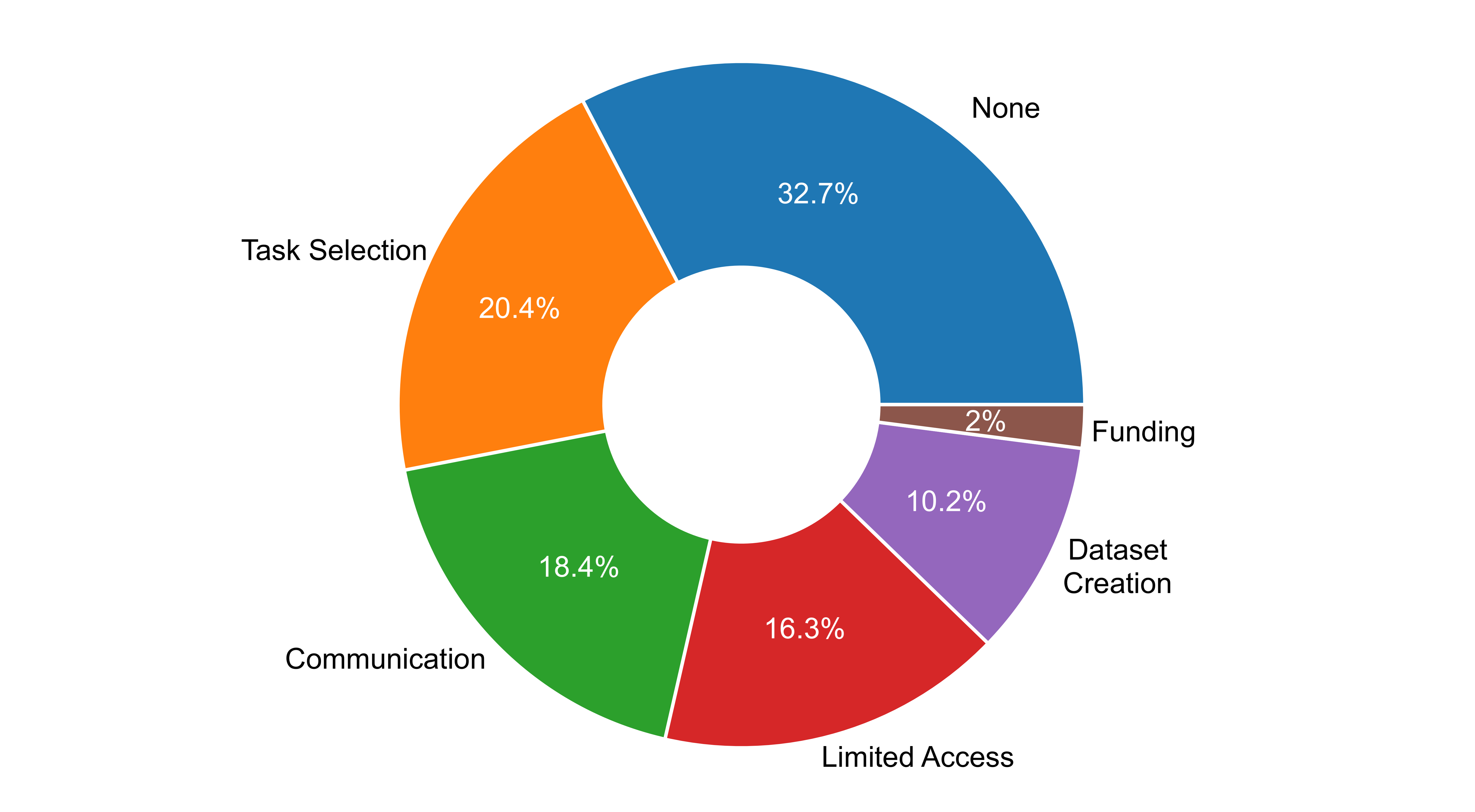}
    \caption{Distribution of %
    the measures participant take to limit misuse of the research artefacts they create. %
    }
    \label{fig:prevent-misuse}
\end{figure}

The next part of the survey sought to uncover how participants currently mitigate misuse of their research artefacts which we inductively code and aggregate (see \Cref{fig:prevent-misuse} for the distribution of coded answers).
The largest share of participants (31.8\%) stated that they do not take any steps to prevent the misuse of their work.
This stands in %
contrast to participants assessment of the risks of misuse of NLP and the tasks they work on, respectively (see \Cref{sec:task-mapping}).
Participants who do seek to mitigate misuse primarily (22.7\%) do so through selecting tasks they believe are not prone to harmful use.
For instance, they highlight factors that they believe are indicative of tasks that could be less prone to dual use, e.g., \textit{"niche tasks which (hopefully) can't be generalised such that they become harmful"}.
However, the particular methods for selecting a task which is not prone to dual use, or one that carries risks of it, are not described, indicating a need for a standardised framework to help guide professionals in selecting tasks.
Participants also measure other methods for addressing dual use concerns, e.g., including ethics statements and outlining limitations of the research in the paper.
While the ethical impacts statements have only recently become mandatory at ACL conferences, they support researchers in clearly communicating their concerns and intentions when publishing research.
Participants also highlight other guidelines, such as data statements~\cite{DBLP:journals/tacl/BenderF18} as helpful in outlining the limitations and risks of their research.
Finally, limiting access to scientific artefacts is also used to mitigate misuse. 
Although this strategy can help mitigate misuse, it can have adverse effects on access for legitimate purposes.
\paragraph{\textbf{Institutional Support}}
The final questions focus on institutional support for participants when it comes to questions of societal harm. 
When asked about support from institutions, only 37.5\% of participants feel positive about the support they receive (see \Cref{fig:likert-scales}).
That is, 
the majority of participants do not feel adequately supported by their institution in terms of preventing the misuse of their research.
Selecting between \textit{Ethics boards, Courses, Training Material}, and \textit{None} most participants selected that they had ethical review boards available (see \Cref{fig:likert-scales}).
While 61\% of participants have previously used institutional support, 14\% of participants stated that they had no access to institutional support for dual-use concerns.
Such a lack of support needs to be addressed from all institutional partners, i.e., publication venues, universities, and companies engaged in the production of research artefacts.
Until there are structured efforts to prevent the misuse of research, however, the responsibility lies with the professionals %
themselves.
Therefore, we propose a checklist in \Cref{sec:checklist} to provide a starting point for researchers to address some of the concerns of dual use. %
\section{Prevention of dual use}\label{sec:prior-work}
In prior work, there are a set of approaches that address the problem of mitigating the misuse of research.  
In the following, we present four approaches mentioned in the context of machine learning at large; \textit{forbidden knowledge}, ethics review boards, education of researchers, and guidelines/checklists.

\subsection{Forbidden Knowledge}\label{sec:forbidden-knowledge}
As knowledge holds power, the term \textit{forbidden knowledge} describes scientific fields that are too dangerous to distribute \citep{johnson1996forbidden}.
In the context of machine learning, \citet{10.2307/23027297} propose to develop and establish an ethical framework to address this field's forbidden knowledge and dual use applications. 
They present a framework to assess forbidden knowledge in machine learning research.
They further propose grades of availability for the applications resulting in a research project, i.e., in which way it could be restricted.
The authors point out that it is infeasible to halt the development of machine learning in directions that could lead to negative outcomes, as machine learning \textit{"is in most cases a general purpose or dual use technology, meaning that it has general capabilities, which are applicable to countless varying purposes"} \citep{10.2307/23027297}.

Contrarily, \citet{DBLP:journals/see/MarchantP09} point out the problems with forbidding science, including: the unenforceability of forbidding a direction of research internationally, especially as international agreements are rare; the legislative imprecision in a rapidly evolving field; the unpredictability of which scientific advances may have harmful applications; the outdated nature of laws regulating science; and the potential misuse of such laws for parochial interests.
The authors emphasise the challenges of identifying dual use of a research artefact ahead of time, as the main research outcome of an artefact might be beneficial and therefore poses the question of the likelihood of destructive or malicious applications.\footnote{Similar to our work, they differentiate dual use and \textit{morally objectionable} research directions.}
They argue that instead of forbidding one research direction altogether, regulation should be put in place to forbid the misuse of research.
Another option would be to pursue the research direction but limit the publication of the scientific results or the models themselves, however the regulations should come from inside the communities, i.e., self-regulation. 
Contributing to that discussion, \citet{brundage-etal-2018-malicious-ai, solaiman-2023-genai} discuss strategies for controlled release of AI models as a mitigation method and the associated trade-offs. 
Going a step further, ~\citet{henderson-etal-2023-self-destructing} argue for more technical approaches to supplement structural strategies to mitigate risks. They demonstrate an approach for embedding task blocking within a language model. 
\subsection{Ethics Review Boards}
In the EU, the \textit{Nuremberg Code} set the precedent for the importance of consent in medical research and later the \textit{Declaration of Helsinki} set the standard for ethical medical research in the EU. 
Any project funded by the EU undergoes an ethics review.\footnote{\url{https://ec.europa.eu/research/participants/data/ref/fp7/89888/ethics-for-researchers_en.pdf}}
In the US and other countries, Institutional Review Boards (IRB) are required by law and are appointed to make decisions on ethics with regard to research on human subjects \cite{GRADY20151148}.

\citet{DBLP:journals/corr/abs-2106-11521} propose the creation of an \textit{Ethics and Society Review board} (ESR), which intervenes at the stage of the submission of research proposals. They argue that currently established ethics boards are concerned with harm to human subjects rather than impact on society at large. 
One of the topics that they focus on is dual use, which they define as \textit{"Any risks or concerns that arise due to the technology being co-opted for nefarious purposes or by motivated actors (e.g., an authoritarian government employed mass surveillance methods)"} \cite{DBLP:journals/corr/abs-2106-11521}.
They test their proposed ESR board and find that the resulting proposals are influenced by the ESR board's focus on ethics. 
Within machine learning, IRB or ERB reviews are typically done for projects involving human annotators or subjects but the wider societal impacts of research are rarely reviewed. 
Recently, pre-registration has been proposed as a method to improve robustness and reliability of machine learning research \citep{van-miltenburg-etal-2021-preregistering, soegaard-etal-2023-two, prereg2021, prereg2022}. While improving scientific rigour, pre-registration could also serve as an opportunity to evaluate the ethical and dual use implications of projects at the proposal stage.

\subsection{Education}
Another path to advocating for a more careful interaction with dual use is the possibility of educating researchers about its danger.
\citet{dualuseeducation} argue for the education of life scientists w.r.t. dual use, starting as part of their university education.
This shifts the responsibility towards universities to provide these courses as well as on to the researchers to have been provided with the courses.
While integrating the topic of dual use in the curriculum is an important issue, it requires a large, coordinated effort, that is yet to materialise. 

With regards to NLP, there have been proposals to integrate ethics into the NLP curriculum, including dual use \citep{DBLP:conf/acl/BenderHS20, strube2022ethics}. In their proposal, \citet{DBLP:conf/acl/BenderHS20} suggest education on dual use by including \textit{"Learning how to anticipate how a developed technology could be repurposed for harmful or negative results, and designing systems so that they do not inadvertently cause harm."}.
We argue that educating researchers about the potential harms of their research should be a core part of their training. 
However, as a prevention method, since it is limited to people receiving formal research training and does not reach all practitioners creating NLP tools, it should be used in conjunction with other methods for prevention of dual use.
\subsection{Checklists and Guidelines}
A proposal presented in previous studies is the implementation of checklists and guidelines for researchers to navigate the complex issue of ethics in their research area~\cite{madaio-et-al-2020-codesigning-checklist, DBLP:conf/emnlp/RogersBL21}.
A wide range of checklists have been created in recent years to cover the topic of ethics and fairness in AI, e.g., ARR's Responsible NLP Research checklist.\footnote{\url{https://aclrollingreview.org/responsibleNLPresearch/}}, NeurIPS' paper checklist\footnote{\url{https://neurips.cc/Conferences/2021/PaperInformation/PaperChecklist}}, and Microsofts' AI Fairness Checklist\footnote{\url{https://www.microsoft.com/en-us/research/project/ai-fairness-checklist/}}
However, existing checklists for AI Ethics at large, and for NLP in specific, do not yet cover the topic of dual use.
Guidelines such as data statements \cite{DBLP:journals/tacl/BenderF18} and model cards \cite{DBLP:conf/fat/MitchellWZBVHSR19} encourage researchers to report on their scientific artefacts more thoroughly, document their intended use, and reflect on the impact those can have.
\citet{DBLP:conf/acl/Mohammad22} present an ethics sheet for AI, which considers the impact and ethics of AI projects. 
Although dual use is briefly mentioned, it does not cover the different aspects of dual use that are required in order to suffice as a starting point for discussion of researchers in the field.
In contrast, the checklist proposed in \Cref{sec:checklist} is designed to be concise and specifically focused on dual use, rather than having long-form text answers as in the ethics sheet.

Checklists as a method in AI ethics have been recently criticised due to their representation of modularity of software and the shifting of responsibility away from the creator of the software \cite{widder2023dislocated}. 
While we attempt to integrate that criticism into the creation of the checklist itself, checklists can only ever be a starting point for a conversation around AI ethics at large and misuse of research artefacts in specific.
Checklists provide a productive first step to help researchers, who are unfamiliar with an area, to approach and consider it. 
Therefore, checklists are a productive starting point and should not be the end-goal.
\section{Checklist for Dual Use}\label{sec:checklist} %
Based on the results of the survey and previously established guidelines and checklists, we propose the following checklist for the issue of dual use in NLP. This checklist should be integrated into existing checklists and guidelines to extend them towards this topic and give researchers a guideline for the issue of malicious reuse of their work.
The design of our checklist is inspired by \citet{DBLP:conf/chi/MadaioSVW20}, who co-design a checklist for fairness in AI with the input of practitioners. 
They argue that checklists should formalise existing informal processes. 
Additionally, our design takes into account the insights of \citet{rosen2010checklist}, who maintain that checklists are essential in complex and specialised fields to ensure that the minimum necessary steps are taken. 
Based on their experience in medicine, \citet{rosen2010checklist} found that simple checklists are most effective in reducing human error. 

The \textit{dual use of scientific artefacts} part of the checklist can be used at the ideation %
stage of a research project as well as in the stages before the publication of the paper. 
We believe that integrating these questions in existing checklists, such as ACL's \textit{Responsible NLP Research Checklist} will draw attention to the topic of dual use and can start a much-needed discussion in the field of NLP on such issues.
\subsection{Checklist Creation}
After collecting the feedback of participants on their perspective of dual use in the survey (\Cref{sec:survey}), we grouped and sorted the feedback to be able to extract questions for the checklist that tackles the issues mentioned by participants. 
We grouped the checklist along the two most important angles for the prevention of misuse as discussed in \Cref{sec:survey-prevention}: the \textit{dual use of scientific artefacts} and \textit{prevention of dual use}.
The \textit{dual use of scientific artefacts} part aims to give researchers space to reflect on their work and the potential for misuse. 
The \textit{prevention of dual use} part aims to make preventive measures explicit and open a discussion about institutional support for the topic of dual use.
\paragraph{\textbf{Dual use of scientific artefacts}}
Despite potential unfamiliarity with the concept of dual use among survey participants, we found that they could identify a range of potential harms deriving from misuse of their work (see \Cref{sec:survey-perception-harm}).
The objective of the first question in the checklist (C1) is to elicit the full range of concerns held by a researcher and ensure they are fully expressed in their paper.
The following questions draw from the potential harms survey participants mentioned in \Cref{sec:survey-perception-harm} as displayed in \Cref{fig:harms}. 
We selected from these harms the ones that can be applied to a wide range, leaving out ethics washing (as it is specific to a subset of areas in NLP), crime (as it is too wide of a question and can be covered by the other topics), cyber bullying (as it could be repetitive with oppression), and plagiarism (as it is again covering only a specific field, potentially repretative). 
This leaves questions about whether scientific artefacts resulting from the research can be used for surveillance (C2), military applications (C3), harm or oppression of (marginalised) groups in society (C4), or to manipulate people by spreading misinformation or polarising users (C5).
\paragraph{\textbf{Prevention of dual use}}
A large share of survey participants stated that they did nothing to prevent misuse of their work (see \Cref{sec:survey-prevention}). 
Therefore, for the second part of the checklist, we focus on ways to mitigate harm through dual use. 
The first question in this section (C6) is aimed to cover all possible ways that participants try to prevent harm through dual use. 
This could be by using institutional support in form of data distribution frameworks or by using licensing options as provided and described by external entities. 
With the following two questions, we focus on institutional support. We found that while a majority of researchers have and access institutional support (see \Cref{sec:survey-prevention}), a large share of participants are yet to use their institutional support. 
In \Cref{sec:prior-work}, we identify a number of possibilities to mitigate the malicious reuse of research that can only be implemented on an institutional level, e.g., ethics review boards and education.
Towards the goal of better education, we phrase the checklist question C7 about ethics training. 
\citet{DBLP:journals/corr/abs-2106-11521} propose the extension of ethics review boards to \textit{Ethics and Society Review board}, indicating the importance of institutional support for questions of societal harm. The checklist question C8 covers these concerns, asking whether the scientific artefacts produced were reviewed for dual use.
\subsection{Checklist}
\textbf{Dual use of scientific artefacts}
\begin{itemize}[noitemsep]
    \item[C1] \textit{Did you explicitly outline the intended use of scientific artefacts you create?}
    \item[C2] \textit{Can any scientific artefacts you create be used for surveillance by companies or governmental institutions?} 
    \item[C3] \textit{Can any scientific artefacts you create be used for military application?}
    \item[C4] \textit{Can any scientific artefacts you create be used to harm or oppress any and particularly marginalised groups of society?}
    \item[C5] \textit{Can any scientific artefacts you create be used to intentionally manipulate, such as spread disinformation or polarise people?}
\end{itemize}

\textbf{Prevention of dual use}
\begin{itemize}[noitemsep]
    \item[C6] \textit{Did you access your institution's or other available resources to ensure limiting the misuse of your research?}
    \item[C7] \textit{Have you been provided by your institution with ethics training that covered potential misuse of your research?} 
    \item[C8] \textit{Were the scientific artefacts you created reviewed for dual use and approved by your institution's ethics board?}
\end{itemize}
\section{Conclusion}
In this paper, we discussed the topic of dual use, the intentional, harmful misuse of research artefacts, in the context of NLP research. 
Dual use has been overlooked in the field of NLP, a gap this paper aims to close.
In order to gauge the current level of consideration for dual use in the field of NLP, we conducted a survey among NLP professionals.
Our survey results revealed that a majority of the participants expressed concerns regarding the potential misuse of their work. They identified several potential dangers associated with their work, including the oppression of (marginalised) groups in society, surveillance, and manipulation. However, they reported taking limited steps toward preventing such misuse.
Based on the participants' inputs on a provided definition of dual use, we propose a definition of dual use that is more appropriate for researchers in the field of NLP.
We discuss existing proposals to address dual use in machine learning and other research fields. Many of these approaches need institutional structures to be realised on a large scale.
Despite the participants' concerns about the potential for misuse of their work in NLP, there appears to be a lack of institutional support for addressing these concerns fully. It is imperative for structured and comprehensive initiatives to be implemented to address this issue.
Utilising the participants' insights, we have developed a dual use checklist for NLP professionals. %
This checklist, which can be integrated with existing ethics checklists, addresses both the potential dual use of scientific artefacts and measures to prevent such dual use.
In conclusion, the dual use of NLP artefacts is a critical concern that has received limited attention in the field.
Our paper strives to raise awareness and initiate a conversation among NLP professionals %
to address this important and overlooked issue.
\section*{Limitations}\label{sec:limitations}
Every study relying on surveys and a mixed-methods approach will have a set of limitations. 
In the case of our study, we find that the participants of the survey underlie a bias based on the location and network of the leading researchers of this project. 
While the survey was distributed both through the researchers' Twitter accounts as well as Reddit, which could potentially reach a more diverse community, most answers are from researchers based in North America and Europe (see \Cref{sec:survey-demographics}). 
This constrains our findings to the issue of dual use as seen through a Western lens, as concerns of dual use may be different across different geographies and social configurations \cite{revill2012lessons}.
It is therefore important to bear in mind that our findings and proposed methods are not necessarily applicable universally and studies that examine the perspective on dual use in other geographies are important for the field of NLP to be able to develop culturally competent methods for addressing dual use.
Further, given the topic and name of the survey, it is likely that participants who are interested in the topic of misuse of research are more likely to answer the full survey. 
However, we do find some researchers who are clearly opposing the underlying premises of the study. 
Quoting one of the comments at the end of the survey, one participant wrote: \textit{"Please don't ruin science."} pointing out their disagreement with the overall topic of ethics in NLP.

\section*{Acknowledgements}
This research was partially funded by a DFF Sapere Aude research leader grant under grant agreement No 0171-00034B, as well as supported by the Pioneer Centre for AI, DNRF grant number P1.

\bibliography{anthology,custom}

\begin{thebibliography}{45}
\expandafter\ifx\csname natexlab\endcsname\relax\def\natexlab#1{#1}\fi

\bibitem[{Albanie et~al.(2021)Albanie, Henriques, Bertinetto,
  Hern{\'a}ndez-Garc{\'\i}a, Doughty, and Varol}]{prereg2022}
Samuel Albanie, Jo\~{a}o~F. Henriques, Luca Bertinetto, Alex
  Hern{\'a}ndez-Garc{\'\i}a, Hazel Doughty, and G\"{u}l Varol, editors. 2021.
\newblock \emph{NeurIPS 2021 Workshop on Pre-registration in Machine Learning},
  volume 181 of \emph{Proceedings of Machine Learning Research}. {PMLR}.

\bibitem[{Bender and Friedman(2018)}]{DBLP:journals/tacl/BenderF18}
Emily~M. Bender and Batya Friedman. 2018.
\newblock \href {https://doi.org/10.1162/tacl\_a\_00041} {Data statements for
  natural language processing: Toward mitigating system bias and enabling
  better science}.
\newblock \emph{Trans. Assoc. Comput. Linguistics}, 6:587--604.

\bibitem[{Bender et~al.(2020)Bender, Hovy, and
  Schofield}]{DBLP:conf/acl/BenderHS20}
Emily~M. Bender, Dirk Hovy, and Alexandra Schofield. 2020.
\newblock \href {https://doi.org/10.18653/v1/2020.acl-tutorials.2} {Integrating
  ethics into the {NLP} curriculum}.
\newblock In \emph{Proceedings of the 58th Annual Meeting of the Association
  for Computational Linguistics: Tutorial Abstracts, {ACL} 2020, Online, July
  5, 2020}, pages 6--9. Association for Computational Linguistics.

\bibitem[{Bentham(1996)}]{bentham1996collected}
Jeremy Bentham. 1996.
\newblock \emph{The collected works of Jeremy Bentham: An introduction to the
  principles of morals and legislation}.
\newblock Clarendon Press.

\bibitem[{Bernstein et~al.(2021)Bernstein, Levi, Magnus, Rajala, Satz, and
  Waeiss}]{DBLP:journals/corr/abs-2106-11521}
Michael~S. Bernstein, Margaret Levi, David Magnus, Betsy Rajala, Debra Satz,
  and Charla Waeiss. 2021.
\newblock \href {http://arxiv.org/abs/2106.11521} {{ESR:} ethics and society
  review of artificial intelligence research}.
\newblock \emph{CoRR}, abs/2106.11521.

\bibitem[{Bertinetto et~al.(2020)Bertinetto, Henriques, Albanie, Paganini, and
  Varol}]{prereg2021}
Luca Bertinetto, Jo{\~a}o~F. Henriques, Samuel Albanie, Michela Paganini, and
  G{\"u}l Varol, editors. 2020.
\newblock \emph{NeurIPS 2020 Workshop on Pre-registration in Machine Learning},
  volume 148 of \emph{Proceedings of Machine Learning Research}. PMLR.

\bibitem[{Biron(2023)}]{Biron_Online_2023}
Bethany Biron. 2023.
\newblock \href
  {https://www.businessinsider.com/company-using-chatgpt-mental-health-support-ethical-issues-2023-1}
  {Online mental health company uses {{ChatGPT}} to help respond to users in
  experiment \textemdash{} raising ethical concerns around healthcare and
  {{AI}} technology}.
\newblock \emph{Business Insider}.

\bibitem[{Brundage et~al.(2018)Brundage, Avin, Clark, Toner, Eckersley,
  Garfinkel, Dafoe, Scharre, Zeitzoff, Filar, Anderson, Roff, Allen,
  Steinhardt, Flynn, h{\'{E}}igeartaigh, Beard, Belfield, Farquhar, Lyle,
  Crootof, Evans, Page, Bryson, Yampolskiy, and
  Amodei}]{brundage-etal-2018-malicious-ai}
Miles Brundage, Shahar Avin, Jack Clark, Helen Toner, Peter Eckersley, Ben
  Garfinkel, Allan Dafoe, Paul Scharre, Thomas Zeitzoff, Bobby Filar, Hyrum~S.
  Anderson, Heather Roff, Gregory~C. Allen, Jacob Steinhardt, Carrick Flynn,
  Se{\'{a}}n~{\'{O}} h{\'{E}}igeartaigh, Simon Beard, Haydn Belfield, Sebastian
  Farquhar, Clare Lyle, Rebecca Crootof, Owain Evans, Michael Page, Joanna
  Bryson, Roman Yampolskiy, and Dario Amodei. 2018.
\newblock \href {http://arxiv.org/abs/1802.07228} {The malicious use of
  artificial intelligence: Forecasting, prevention, and mitigation}.
\newblock \emph{CoRR}, abs/1802.07228.

\bibitem[{EU(2023)}]{EU_proposal_2023}
EU. 2023.
\newblock \href
  {https://eur-lex.europa.eu/legal-content/EN/TXT/?uri=CELEX:52021PC0206}
  {Artificial intelligence act: Proposal for a regulation of the european
  parliament and of the council laying down harmonised rules on artificial
  intelligence act and amending certain union legislative acts}.

\bibitem[{Forge(2010)}]{forge2010note}
John Forge. 2010.
\newblock A note on the definition of “dual use”.
\newblock \emph{Science and Engineering Ethics}, 16(1):111--118.

\bibitem[{Foucault(2012)}]{foucault2012discipline}
Michel Foucault. 2012.
\newblock \emph{Discipline and punish: The birth of the prison}.
\newblock Vintage.

\bibitem[{Gamage et~al.(2021)Gamage, Sasahara, and
  Chen}]{DBLP:conf/tto/GamageSC21}
Dilrukshi Gamage, Kazutoshi Sasahara, and Jiayu Chen. 2021.
\newblock \href
  {https://truthandtrustonline.com/wp-content/uploads/2021/10/TTO2021\_paper\_20.pdf}
  {The emergence of deepfakes and its societal implications: {A} systematic
  review}.
\newblock In \emph{Proceedings of the 2021 Truth and Trust Online Conference
  {(TTO} 2021), Virtual, October 7-8, 2021}, pages 28--39. Hacks Hackers.

\bibitem[{Goldstein et~al.(2023)Goldstein, Sastry, Musser, DiResta, Gentzel,
  and Sedova}]{DBLP:journals/corr/abs-2301-04246}
Josh~A. Goldstein, Girish Sastry, Micah Musser, Renee DiResta, Matthew Gentzel,
  and Katerina Sedova. 2023.
\newblock \href {https://doi.org/10.48550/arXiv.2301.04246} {Generative
  language models and automated influence operations: Emerging threats and
  potential mitigations}.
\newblock \emph{CoRR}, abs/2301.04246.

\bibitem[{Grady(2015)}]{GRADY20151148}
Christine Grady. 2015.
\newblock \href {https://doi.org/https://doi.org/10.1378/chest.15-0706}
  {Institutional review boards: Purpose and challenges}.
\newblock \emph{Chest}, 148(5):1148--1155.

\bibitem[{Henderson et~al.(2023)Henderson, Mitchell, Manning, Jurafsky, and
  Finn}]{henderson-etal-2023-self-destructing}
Peter Henderson, Eric Mitchell, Christopher Manning, Dan Jurafsky, and Chelsea
  Finn. 2023.
\newblock \href {https://doi.org/10.1145/3600211.3604690} {Self-destructing
  models: Increasing the costs of harmful dual uses of foundation models}.
\newblock In \emph{Proceedings of the 2023 AAAI/ACM Conference on AI, Ethics,
  and Society}, AIES '23, page 287–296, New York, NY, USA. Association for
  Computing Machinery.

\bibitem[{Hovy et~al.(2020)Hovy, Bianchi, and
  Fornaciari}]{hovy-etal-2020-sound}
Dirk Hovy, Federico Bianchi, and Tommaso Fornaciari. 2020.
\newblock \href {https://doi.org/10.18653/v1/2020.acl-main.154} {{``}you sound
  just like your father{''} commercial machine translation systems include
  stylistic biases}.
\newblock In \emph{Proceedings of the 58th Annual Meeting of the Association
  for Computational Linguistics}, pages 1686--1690, Online. Association for
  Computational Linguistics.

\bibitem[{Hovy and Spruit(2016)}]{DBLP:conf/acl/HovyS16}
Dirk Hovy and Shannon~L. Spruit. 2016.
\newblock \href {https://doi.org/10.18653/v1/p16-2096} {The social impact of
  natural language processing}.
\newblock In \emph{Proceedings of the 54th Annual Meeting of the Association
  for Computational Linguistics, {ACL} 2016, August 7-12, 2016, Berlin,
  Germany, Volume 2: Short Papers}. The Association for Computer Linguistics.

\bibitem[{Johnson(1996)}]{johnson1996forbidden}
Deborah~G Johnson. 1996.
\newblock Forbidden knowledge and science as professional activity.
\newblock \emph{The Monist}, 79(2):197--217.

\bibitem[{Kania(2018)}]{kania2018technological}
Elsa Kania. 2018.
\newblock Technological entanglement: Cooperation, competition and the dual-use
  dilemma in artificial intelligence.

\bibitem[{Kempner et~al.(2011)Kempner, Merz, and Bosk}]{10.2307/23027297}
Joanna Kempner, Jon~F. Merz, and Charles~L. Bosk. 2011.
\newblock \href {http://www.jstor.org/stable/23027297} {Forbidden knowledge:
  Public controversy and the production of nonknowledge}.
\newblock \emph{Sociological Forum}, 26(3):475--500.

\bibitem[{Koplin(2023)}]{koplin_dual-use_2023}
Julian~J. Koplin. 2023.
\newblock \href {https://doi.org/10.1007/s10676-023-09703-z} {Dual-use
  implications of {AI} text generation}.
\newblock \emph{Ethics and Information Technology}, 25(2):32.

\bibitem[{Leins et~al.(2020)Leins, Lau, and Baldwin}]{DBLP:conf/acl/LeinsLB20}
Kobi Leins, Jey~Han Lau, and Timothy Baldwin. 2020.
\newblock \href {https://doi.org/10.18653/v1/2020.acl-main.261} {Give me
  convenience and give her death: Who should decide what uses of {NLP} are
  appropriate, and on what basis?}
\newblock In \emph{Proceedings of the 58th Annual Meeting of the Association
  for Computational Linguistics, {ACL} 2020, Online, July 5-10, 2020}, pages
  2908--2913. Association for Computational Linguistics.

\bibitem[{Madaio et~al.(2020{\natexlab{a}})Madaio, Stark, Vaughan, and
  Wallach}]{DBLP:conf/chi/MadaioSVW20}
Michael~A. Madaio, Luke Stark, Jennifer~Wortman Vaughan, and Hanna~M. Wallach.
  2020{\natexlab{a}}.
\newblock \href {https://doi.org/10.1145/3313831.3376445} {Co-designing
  checklists to understand organizational challenges and opportunities around
  fairness in {AI}}.
\newblock In \emph{{CHI} '20: {CHI} Conference on Human Factors in Computing
  Systems, Honolulu, HI, USA, April 25-30, 2020}, pages 1--14. {ACM}.

\bibitem[{Madaio et~al.(2020{\natexlab{b}})Madaio, Stark, Wortman~Vaughan, and
  Wallach}]{madaio-et-al-2020-codesigning-checklist}
Michael~A. Madaio, Luke Stark, Jennifer Wortman~Vaughan, and Hanna Wallach.
  2020{\natexlab{b}}.
\newblock \href {https://doi.org/10.1145/3313831.3376445} {Co-designing
  checklists to understand organizational challenges and opportunities around
  fairness in ai}.
\newblock In \emph{Proceedings of the 2020 CHI Conference on Human Factors in
  Computing Systems}, CHI '20, page 1–14, New York, NY, USA. Association for
  Computing Machinery.

\bibitem[{Mahfoud et~al.(2018)Mahfoud, Aircardi, Datta, and
  Rose}]{mahfoud-etal-2018}
Tara Mahfoud, Christine Aircardi, Saheli Datta, and Nikolas Rose. 2018.
\newblock \href {https://www.jstor.org/stable/26597992} {The limits of dual
  use}.
\newblock \emph{Issues in Science and Technology}, 34(4):73--78.

\bibitem[{Marchant and Pope(2009)}]{DBLP:journals/see/MarchantP09}
Gary~E. Marchant and Lynda~L. Pope. 2009.
\newblock \href {https://doi.org/10.1007/s11948-009-9130-9} {The problems with
  forbidding science}.
\newblock \emph{Sci. Eng. Ethics}, 15(3):375--394.

\bibitem[{Michael et~al.(2022)Michael, Holtzman, Parrish, Mueller, Wang, Chen,
  Madaan, Nangia, Pang, Phang, and Bowman}]{DBLP:journals/corr/abs-2208-12852}
Julian Michael, Ari Holtzman, Alicia Parrish, Aaron Mueller, Alex Wang,
  Angelica Chen, Divyam Madaan, Nikita Nangia, Richard~Yuanzhe Pang, Jason
  Phang, and Samuel~R. Bowman. 2022.
\newblock \href {https://doi.org/10.48550/arXiv.2208.12852} {What do {NLP}
  researchers believe? results of the {NLP} community metasurvey}.
\newblock \emph{CoRR}, abs/2208.12852.

\bibitem[{Minehata et~al.(2013)Minehata, Sture, Shinomiya, Whitby, and
  Dando}]{dualuseeducation}
Masamichi Minehata, Judi Sture, Nariyoshi Shinomiya, Simon Whitby, and Malcolm
  Dando. 2013.
\newblock \href {https://doi.org/10.20965/jdr.2013.p0674} {Promoting education
  of dual-use issues for life scientists: A comprehensive approach}.
\newblock \emph{Journal of Disaster Research}, 8:674--685.

\bibitem[{Mitchell et~al.(2019)Mitchell, Wu, Zaldivar, Barnes, Vasserman,
  Hutchinson, Spitzer, Raji, and Gebru}]{DBLP:conf/fat/MitchellWZBVHSR19}
Margaret Mitchell, Simone Wu, Andrew Zaldivar, Parker Barnes, Lucy Vasserman,
  Ben Hutchinson, Elena Spitzer, Inioluwa~Deborah Raji, and Timnit Gebru. 2019.
\newblock \href {https://doi.org/10.1145/3287560.3287596} {Model cards for
  model reporting}.
\newblock In \emph{Proceedings of the Conference on Fairness, Accountability,
  and Transparency, FAT* 2019, Atlanta, GA, USA, January 29-31, 2019}, pages
  220--229. {ACM}.

\bibitem[{Mohammad(2022)}]{DBLP:conf/acl/Mohammad22}
Saif~M. Mohammad. 2022.
\newblock \href {https://doi.org/10.18653/v1/2022.acl-long.573} {Ethics sheets
  for {AI} tasks}.
\newblock In \emph{Proceedings of the 60th Annual Meeting of the Association
  for Computational Linguistics (Volume 1: Long Papers), {ACL} 2022, Dublin,
  Ireland, May 22-27, 2022}, pages 8368--8379. Association for Computational
  Linguistics.

\bibitem[{Ratner(2021)}]{ratner2021sweetie}
Claudia Ratner. 2021.
\newblock When “sweetie” is not so sweet: Artificial intelligence and its
  implications for child pornography.
\newblock \emph{Family Court Review}, 59(2):386--401.

\bibitem[{Resnik(2009)}]{DBLP:journals/see/Resnik09}
David~B. Resnik. 2009.
\newblock \href {https://doi.org/10.1007/s11948-008-9104-3} {What is "dual use"
  research? {A} response to miller and selgelid}.
\newblock \emph{Sci. Eng. Ethics}, 15(1):3--5.

\bibitem[{Revill et~al.(2012)Revill, Carnevali, Forsberg, Holmstr{\"o}m, Rath,
  Shinwari, and Mancini}]{revill2012lessons}
James Revill, M~Daniela~Candia Carnevali, {\AA}ke Forsberg, Anna Holmstr{\"o}m,
  Johannes Rath, Zabta~Khan Shinwari, and Giulio~M Mancini. 2012.
\newblock Lessons learned from implementing education on dual-use in austria,
  italy, pakistan and sweden.
\newblock \emph{Medicine, Conflict and Survival}, 28(1):31--44.

\bibitem[{Rogers et~al.(2021)Rogers, Baldwin, and
  Leins}]{DBLP:conf/emnlp/RogersBL21}
Anna Rogers, Timothy Baldwin, and Kobi Leins. 2021.
\newblock \href {https://doi.org/10.18653/v1/2021.findings-emnlp.414} {'just
  what do you think you're doing, dave?' {A} checklist for responsible data use
  in {NLP}}.
\newblock In \emph{Findings of the Association for Computational Linguistics:
  {EMNLP} 2021, Virtual Event / Punta Cana, Dominican Republic, 16-20 November,
  2021}, pages 4821--4833. Association for Computational Linguistics.

\bibitem[{Rosen(2010)}]{rosen2010checklist}
Dennis Rosen. 2010.
\newblock The checklist manifesto: How to get things right.
\newblock \emph{JAMA}, 303(7):670--673.

\bibitem[{Schmid et~al.(2022)Schmid, Riebe, and
  Reuter}]{DBLP:journals/see/SchmidRR22}
Stefka Schmid, Thea Riebe, and Christian Reuter. 2022.
\newblock \href {https://doi.org/10.1007/s11948-022-00364-7} {Dual-use and
  trustworthy? {A} mixed methods analysis of {AI} diffusion between civilian
  and defense r{\&}d}.
\newblock \emph{Sci. Eng. Ethics}, 28(2):12.

\bibitem[{Shankar and Zare(2022)}]{DBLP:journals/natmi/ShankarZ22}
Sadasivan Shankar and Richard~N. Zare. 2022.
\newblock \href {https://doi.org/10.1038/s42256-022-00481-9} {The perils of
  machine learning in designing new chemicals and materials}.
\newblock \emph{Nat. Mach. Intell.}, 4(4):314--315.

\bibitem[{S{\o}gaard et~al.(2023)S{\o}gaard, Hershcovich, and
  de~Lhoneux}]{soegaard-etal-2023-two}
Anders S{\o}gaard, Daniel Hershcovich, and Miryam de~Lhoneux. 2023.
\newblock \href {https://doi.org/10.18653/v1/2023.eacl-main.6} {A two-sided
  discussion of preregistration of {NLP} research}.
\newblock In \emph{Proceedings of the 17th Conference of the European Chapter
  of the Association for Computational Linguistics}, pages 83--93, Dubrovnik,
  Croatia. Association for Computational Linguistics.

\bibitem[{Solaiman(2023)}]{solaiman-2023-genai}
Irene Solaiman. 2023.
\newblock \href {https://doi.org/10.1145/3593013.3593981} {The gradient of
  generative ai release: Methods and considerations}.
\newblock In \emph{Proceedings of the 2023 ACM Conference on Fairness,
  Accountability, and Transparency}, FAccT '23, page 111–122, New York, NY,
  USA. Association for Computing Machinery.

\bibitem[{Strube and Pan(2022)}]{strube2022ethics}
Michael Strube and Shimei Pan. 2022.
\newblock Ethics in nlp: Bias and dual use.
\newblock \emph{UMBC Faculty Collection}.

\bibitem[{Urbina et~al.(2022)Urbina, Lentzos, Invernizzi, and
  Ekins}]{DBLP:journals/natmi/UrbinaLIE22}
Fabio Urbina, Filippa Lentzos, C{\'{e}}dric Invernizzi, and Sean Ekins. 2022.
\newblock \href {https://doi.org/10.1038/s42256-022-00465-9} {Dual use of
  artificial-intelligence-powered drug discovery}.
\newblock \emph{Nat. Mach. Intell.}, 4(3):189--191.

\bibitem[{van Miltenburg et~al.(2021)van Miltenburg, van~der Lee, and
  Krahmer}]{van-miltenburg-etal-2021-preregistering}
Emiel van Miltenburg, Chris van~der Lee, and Emiel Krahmer. 2021.
\newblock \href {https://doi.org/10.18653/v1/2021.naacl-main.51}
  {Preregistering {NLP} research}.
\newblock In \emph{Proceedings of the 2021 Conference of the North American
  Chapter of the Association for Computational Linguistics: Human Language
  Technologies}, pages 613--623, Online. Association for Computational
  Linguistics.

\bibitem[{Vanmassenhove et~al.(2018)Vanmassenhove, Hardmeier, and
  Way}]{vanmassenhove2018getting}
Eva Vanmassenhove, Christian Hardmeier, and Andy Way. 2018.
\newblock \href {https://doi.org/10.18653/v1/D18-1334} {Getting gender right in
  neural machine translation}.
\newblock In \emph{Proceedings of the 2018 Conference on Empirical Methods in
  Natural Language Processing}, pages 3003--3008, Brussels, Belgium.
  Association for Computational Linguistics.

\bibitem[{Weidinger et~al.(2021)Weidinger, Mellor, Rauh, Griffin, Uesato,
  Huang, Cheng, Glaese, Balle, Kasirzadeh, Kenton, Brown, Hawkins, Stepleton,
  Biles, Birhane, Haas, Rimell, Hendricks, Isaac, Legassick, Irving, and
  Gabriel}]{DBLP:journals/corr/abs-2112-04359}
Laura Weidinger, John Mellor, Maribeth Rauh, Conor Griffin, Jonathan Uesato,
  Po{-}Sen Huang, Myra Cheng, Mia Glaese, Borja Balle, Atoosa Kasirzadeh, Zac
  Kenton, Sasha Brown, Will Hawkins, Tom Stepleton, Courtney Biles, Abeba
  Birhane, Julia Haas, Laura Rimell, Lisa~Anne Hendricks, William~S. Isaac,
  Sean Legassick, Geoffrey Irving, and Iason Gabriel. 2021.
\newblock \href {http://arxiv.org/abs/2112.04359} {Ethical and social risks of
  harm from language models}.
\newblock \emph{CoRR}, abs/2112.04359.

\bibitem[{Widder and Nafus(2023)}]{widder2023dislocated}
David~Gray Widder and Dawn Nafus. 2023.
\newblock Dislocated accountabilities in the “{AI} supply chain”:
  {M}odularity and developers’ notions of responsibility.
\newblock \emph{Big Data \& Society}, 10(1):20539517231177620.

\end{thebibliography}
\bibliographystyle{acl_natbib}
\newpage
\appendix
\section{Codebooks}
For each of the free-form answer fields of the survey, we coded the answers of the participants, for details see Section~\ref{sec:survey}. Below we provide the codebooks for the relevant questions reported in this paper.

\begin{table*}[!ht]
    \centering
    \begin{tabular}{{p{25mm}p{30mm}p{30mm}p{30mm}}}
    \toprule
        code & description & includes(sub-codes) & examples \\ \midrule
        surveillance & automation of surveillance by any entity & surveillance by corporations; surveillance by government & large scale data collection and processing of users \\ \midrule 
        manipulation & manipulation of users of technology, trying to change their beliefs or world-view & disinformation, polarisation & disinformation generation \\ \midrule
        oppression of groups in society & technologies to oppress, marginalise or disadvantage any group in a society & language standardisation; propagation of racist believes or classification & generation of hate speech; classification based on dialects in a language \\ \midrule
        crime & reuse of technologies for general criminal purpose & ~ & personalised phishing \\ \midrule
        ethics washing & reuse technologies to minimise the underlying problems of a system & ~ & explainability to fairwash an algorithm exhibiting bias \\ \midrule
        cyber bullying & target individuals with harmful, oppressive or disadvantageous content & ~ & using technologies to bully kids in schools \\ \midrule
        military application & reuse of technologies by the military & ~ & target systems \\ \midrule
        censorship & automation of censorship & censorship by government & finding content to censor on a large scale \\ \midrule
        plagiarism & automatic creation or summarisation of content for plagiarism & ~ & automatically rephrasing existing academic work to republish \\ \bottomrule
    \end{tabular}
    \caption{Codebook for the harms identified by participants for each of the tasks they work on. Question: \textit{Please add for each task the following: Dual Use Use Cases: In the text field, state which use cases of the task could be problematic (e.g., uses when models are intentionally deployed out of context or cause harm)}.}
    \label{tab:codebook-hamrs}
\end{table*}
\subsection{Harms by Task}
Participants were asked to list the potential harm of each of the tasks they stated they worked on. The codebook is provided in Table~\ref{tab:codebook-hamrs}. The distribution of the codes among the survey respondents can be found in Figure~\ref{fig:harms}.

\begin{table*}[!ht]
    \centering
    \begin{tabular}{p{25mm}p{50mm}p{20mm}p{40mm}}
    \toprule
        code & description & includes (sub-codes) & examples \\ \midrule
        Task Selection & considerations regarding the tasks the participants work on & ~ & selecting tasks or research problems to work on that are less prone to misuse; decline jobs or research projects where participants are not comfortable how their models are used \\ \midrule
        Communication & communicating the risks related through the work; discussing potential risks with the community & ethics statements & writing the paper to clearly communicate risks; using ethics statements \\\midrule
        Limited Access & limiting access to scientific artefacts & regulation work; licensing & limit access to data and code \\ \midrule
        Dataset Creation & careful considerations when the dataset is created with regards to how it will be reused & dataset statements & anonymisation of datasets; keep processing of data local\\ \midrule
        Funding & considerations regarding the sources of funding for research & ~ & selection and rejecting projects based on the funding agency \\\midrule
        None & no measures taken to limit malicious reuse & ~ & comment "None" \\ \bottomrule
    \end{tabular}
    \caption{Codebook for how participants select tasks to work on. Question: \textit{What measures do you take to limit the potential misuse of your research (if any)?}.}
    \label{tab:codebook-measures}
\end{table*}
[h]
\subsection{Measures to limit misuse}
Participants were asked to list measures they take to limit the potential misuse of their work. The codebook is provided in Table~\ref{tab:codebook-measures}. The distribution of the codes among the survey respondents can be found in Figure~\ref{fig:prevent-misuse}.

\section{Survey}\label{app:survey-setup}
We provide the full survey as conducted on LimeSurvey below.

\includepdf[pages=-,scale=0.7]{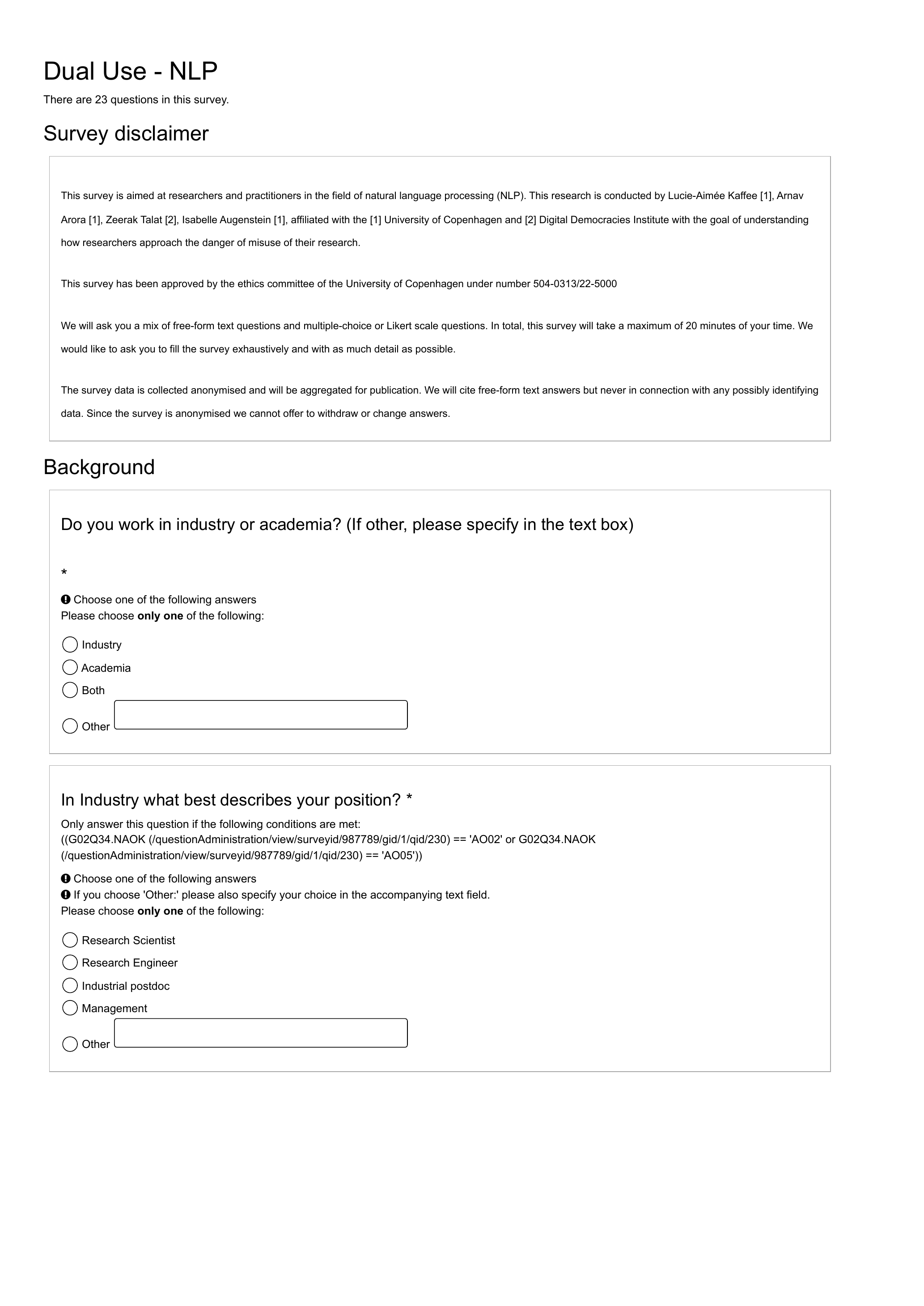}

\end{document}